\documentclass[conference]{IEEEtran}
\IEEEoverridecommandlockouts
\usepackage{cite}
\usepackage{amsmath,amssymb,amsfonts}
\usepackage{algorithmic}
\usepackage{graphicx}
\usepackage{textcomp}
\usepackage{xcolor}
\usepackage{indentfirst}
\usepackage{caption}

\usepackage{hyperref}
\def\BibTeX{{\rm B\kern-.05em{\sc i\kern-.025em b}\kern-.08em
    T\kern-.1667em\lower.7ex\hbox{E}\kern-.125emX}}

\begin{document}

\title{Robotic Exploration for Mapping\\
}

\author{\IEEEauthorblockN{ Akanimoh 'Sanmi Adeleye}
\IEEEauthorblockA{\textit{University of California, San Diego} \\
\textit{Contextual Robotics Institute}\\
 akadeley@eng.ucsd.edu}
}

\maketitle

\begin{abstract}
 Robotic Exploration has evolved rapidly in the past two decades as new and more complex techniques have been created to explore unknown regions efficiently. Exciting advancements in exploration, autonomous navigation, and sensor technology have created opportunities for robots to be utilized in new environments and for new objectives ranging from mapping of abandon mines and deep oceans to the efficient creation of indoor models for navigation and search. In this paper we present and discuss a number of examples in research literature of these recent advancements, specifically focusing on robotic exploration algorithms for unmanned vehicles.
\end{abstract}


\section{Introduction}
In general, exploration can simply be defined as actively acquiring previously unknown information about the world; to achieve an objective. In robotics, the objective is often to gain complete information about the latent spatial/geometric layout of the world. The robot must decide, based on the current understanding of the latent space and sensor information, what action to take to complete the objective in a reasonable amount of time. Most ground exploration robots rely on one or two types of perception sensors to collect data: laser scanners and RGB cameras \cite{choudhary2018autonomous}. This leads to numerous different environments and objectives an exploration mission can have. 

Autonomous and teleoperated robotic exploration has been applied to all domains of exploration. This survey focuses on autonomous robotic exploration. Autonomous exploration is foremost used in cases where remote control is impractical, such as planetary exploration, and/or there is limited or no communication links to the robots. It is also useful in environments where autonomy simplifies the objective or reduces the level of danger for human works and operators. For this reason, autonomous exploration can be more beneficial than teleoperation even within environments where remote control is possible. This is true for all domains but highlighted within the underwater, were although teleoperation is used particularly heavily, autonomous exploration can provide operator fatigue relief from extended periods of operation and repetitive, tedious task. 

The goal of this work is to give readers an overview of the core exploration methods used within the robotic community, the motivation behind them, and divergences between them. In section \hyperref[sec:Problem Formulation]{II} we begin by formalizing the three different primary methods for exploration. Section \hyperref[sec:MapRep]{III} gives a brief overview of the different types of map representations that exploration methods use while exploring and present as a final model of the environment. Section \hyperref[sec:ExpMethod]{IV} reviews advancements in the research literature and categorizes the different approaches. In section \hyperref[sec:ExpPC]{V} we provide a discussion of exploration method and environments they are used in, and in section \hyperref[sec:DiscussionOP]{VI} we present open problems in the field. Finally, section \hyperref[sec:Conclusion]{VII} presents our conclusions.

\section{Problem Formulation}
\label{sec:Problem Formulation}
Exploration is a decision making problem and therefore can most generally be formulated as a POMDP. Due to the inherent computational intractability of POMDPs, most approaches reformulate this problem using an Information theoretic, Greedy, or Learning-based model. In this section we first review the POMDP formulation and explain why it becomes intractable, then formulate the three other most common approaches.

\subsection{Exploration as a POMDP} 
Partially observable Markov decision processes (POMDP)\cite{kaelbling1998planning} are an extension of MDP where complete state information is not given but rather observed with some noise. In this way, they are more representative of how real robots interact with the world. Similar to MDPs, POMDPs have finite states $S$ and actions $A$, a transition function $T$ and a reward $R$. They also have a policy $\pi(s) = a$, which describes the behavior of an agent given a state. Additionally, they include a finite set of observations $\Omega$ that hold information about the preceding action and current state through a noisy observation function $O:S \times A \rightarrow \Pi(\Omega)$. Which is written as $O(s',a,o)$: the probability of observing $o$ given action $a$ and arriving in the current state $s'$. The goal is to find a policy that maximizes the future cumulative reward.

Consider an initial belief state distribution $b_0(s)= Pr(s_0 = s)$. This is the probability the agent assigns to world state $s$ by belief $b$. At each time step, the agent can use Bayesian reasoning to generate an updated belief based on the prior belief, action, and current observation: 
\begin{equation}
\begin{aligned}
\label{eq:1}
       b_{t+1}(s') = Pr(s_{t+1} = s'  | b_{t} = b, a_{t} = a, o_{t+1} = o) 
\end{aligned}
\end{equation}

Because this belief state is a sufficient statistic, the agents policy can be conditioned on this current belief rather than the unknown state distribution: $a_t = \pi(b_t)$. This change of variables gives rise to the "belief" MDP, where the transition function and reward function are reformulated to include beliefs over states. 

In general, for most exploration problems we are concerned with reaching/understanding a given state with high certainty (finding an optimal policy that reduces uncertainty). The optimal policy within the belief model (in the discounted infinite horizon case) will lead to the optimal policy for the POMDP. The pose of the robot in the real world can be defined by six dimensions, while the state and observation space grow depending on the environment and number of sensors available to the robot. In both the belief and state space, finding the posterior distribution needed becomes intractable do to the high dimensionality of the action, state, and observation space. A common approach is to discrete the state and observation space or, in the belief space, perform approximations using appropriate cost functions (typically through some sort of Gaussian process) \cite{platt2010belief,martinez2009bayesian}. Information gain is one example that has been shown to be a useful theoretical method to design such a function and is formalized in the next section.

\subsection{Information Gain}
\label{sec:InfoGain}
In this section we will first formalize the exploration problem using an information gain approach. See section \hyperref[sec:InfoGain]{IV.B} for application based approaches using this formalization. 

 In the context of exploration, entropy $H_p(x)$ is defined as the expected information of the log of a probability distribution: $E[-log p(x)]$.\cite{thrun2005probabilistic}  
\begin{equation}
\begin{aligned}
\label{eq:2}
       H_{p}(x) = -\int p(x) \log p(x) dx  \text{ or } -\sum p(x) \log p(x) 
\end{aligned}
\end{equation}

As mentioned in the section above, for exploration, we are generally concerned with reaching a given state with high certainty (often as quickly as possible). Our belief over the state distribution is how the robot internally understands the world and determines the probability of a given state. The objective is to find optimal actions that reduce uncertainty in the posterior probability (Eq. \ref{eq:1}). The entropy of a belief state, conditioned on possible actions and expected observation, is a measure of uncertainty in the belief and therefore can be used to determine the expected reduction in uncertainty \cite{stachniss2005information,bourgault2002information}. 

\begin{equation}
\begin{aligned}
\label{eq:3}
       H_{b}(x'| z,u) = -\int P(x'|b,z,u) \log P(x'|b,z,u) dx'  
\end{aligned}
\end{equation}

We write Eq. \ref{eq:3} using the variables common in literature where $x$ is the state of the robot, typically the pose and map, $z$ is the noisy observations, and $u$ is the control signal for the robot to perform an action. The difference between the entropy of the current state and an expected future state, is called the \textit{information gain}. Evaluated over all possible actions, we can determine which action leads to a future state that is expected to reduce uncertainty the most. 

\begin{equation}
\begin{aligned}
\label{eq:4}
       I_{b}(u) = H(x) - E_{z} [ H_{b}(x'|z,u)]   
\end{aligned}
\end{equation}

Because we only have a choice over the control signal and not our observations, to evaluate the expected entropy in the information gain, we integrate Eq. \ref{eq:3} over all possible observations. Where observations are approximated based on map representation and sensor type. 

\begin{equation}
\begin{aligned}
\label{eq:5}
       H_{b}(x'|u) = -\int \int_{z} P(x'|b,z,u) \log P(x'|b,z,u) dz \, dx'  
\end{aligned}
\end{equation}

\subsection{Greedy Techniques}
\label{sec:Greddy}
In this section, we formulate the exploration problem broadly for greedy techniques. Given a belief about the world (Eq. \ref{eq:1}), based on sensor observations, we can partition the map into known and or unknown regions. Known regions are areas that have been covered by the sensors up till now, where as unknown regions have not yet been explored by sensors. The greedy approach tries to eliminate all unknown regions as quickly as possible. 

Formally the greedy technique can be seen as the optimal location to move to given the expected utility ($U$) of that actions minus the cost ($C$) to move there: 

\begin{equation}
\begin{aligned}
\label{eq:6}
     \arg\max_{u_{t}}  E_{x_{t+1},z_{t+1}} [U(b_{t+1}| u_{t},z_{t})] - C(b_{t})
\end{aligned}
\end{equation}

With information gain approaches, it is possible to consider methods that are myopic (particularly a time horizon of T=1) as greedy techniques. This is shown in Eq. \ref{eq:7} where the utility of an action is the expected information gain and cost is defined according to the implementation (most often euclidean distance). 

\begin{equation}
\begin{aligned}
\label{eq:7}
     \arg\max_{u_{t}} H(b_{t}) - E_{x_{t+1},z_{t+1}} [H(b_{t+1}| u_{t},z_{t})] - C(b_{t})
\end{aligned}
\end{equation}

For this survey, we chose to group these methods separately because entropy based approaches inherently tend to optimize the utility of the next action in relation to future states. Instead, our formulation is most fitting for frontier, sampling, and a few other myopic based approaches.

\subsection{Learning-Based Approach} 
\label{sec:Learning}
As mentioned earlier, exploration at its core is a decision theoretical problem where you want to optimization actions given a belief. There are a verity of approaches to discretize the space and/or approximate your belief to optimize action selection. We have shown a greedy approach where an objective function is created by hand but is focused on optimizing the current state. We then explained through an entropy based information gain approach, methods where the objective function is still designed by hand but looks beyond only optimizing the current state, to future states as well. Finally we present optimization techniques where the objective function is created by hand but its meaning and value are learned through experience.  

\begin{figure*} [!ht]
\includegraphics[width=\textwidth,height=7cm]{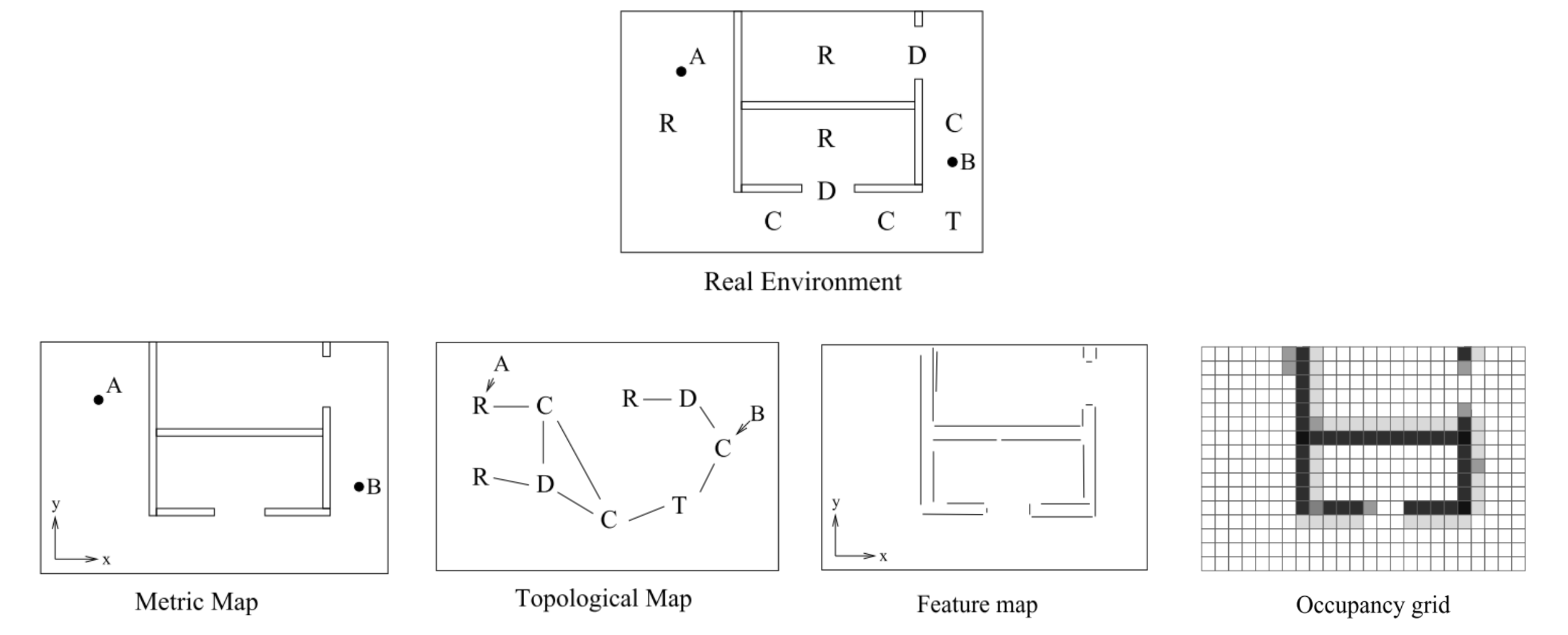}
\label{fig:MapRep}
\caption{The figure above is depicts a real environment represented by four commonly used models. This figure is adapted from \cite{filliat2003map} where R=room, C=corridor, T=turn, D=door. Positions A and B are seen in the metric and topological maps. Within the metric map, the positions are located according to the local reference frame, whereas in the topological map, they are associated with R and C and are inferred to be connected. The feature map, depicts segments of obstacle in the environment and the occupancy grid map discreteness the environment with probabilities of object occupancy (0 for white and 1 for black).}
\end{figure*}

Learning based optimization techniques are categorized by the influence data (experience) has over the objective function. This can be directly, as seen in supervised learning, where the model learns to classify based on data and associated labels: Naive Bayes, Hidden Markov Models(HMM), Linear and Logistic Regression, Support vector mechanics, Neural Networks, etc; as well as more indirectly, through unsupervised and reinforcement learning: Expectation Maximization, K-means, HMM, Neural Networks, Value and Policy iteration, etc. In this section we walk through a Value iteration formulation for exploration, a commonly used approach that aligns well with our previous formulations. 

Value iteration recursively calculates the utility of each action according to the objective function and current belief. This formulation once again grows to be intractable with a large belief space. We formalize this approach with a MDP based value iteration, which is accurate for a discretized state and action space and/or an approximated belief space. The value of a state is initially set to some constant then updated as follows: 

\begin{equation}
\begin{aligned}
\label{eq:8}
     V(x) \leftarrow \gamma \max_{u} [ r(x,u) + \int V(x')p(x'|u,x)dx'] 
\end{aligned}
\end{equation}

The reward function, $r(x,u)$, is designed to be given from the environment, based on progress towards the goal. Here $r(x,u)$ is the expected discounted reward over time. The policy is can then be determined from the current value function:

\begin{equation}
\begin{aligned}
\label{eq:9}
     \pi(x) = \arg\max_{u} [ r(x,u) + \int V(x')p(x'|u,x)dx'] 
\end{aligned}
\end{equation}

\section{Map Representations}
\label{sec:MapRep}
While exploration can be abstracted as shown above, the objective of the robot and map representation, determines the exploration strategy and implementation used. In this section we provide a brief overview of common map representations. We do not go in detail about implementation techniques but rather focus on understanding representation and the differences between them. We refer readers to \cite{durrant2006simultaneous}, \cite{bailey2006simultaneous}, \cite{cadena2016past} and \cite{thrun2002robotic} for more details. 

As an overview, mapping has traditionally been split in two categories: metric and topological. Metric maps captures the geometric properties of the environment, such as position of objects within a common reference frame; topological maps describe the environment by the relative connectivity of distinct significant places. The difference between the two categories is not rigid since practically all topological maps use some geometric information. However, a clear distinction between the two is shown when considering transforms performed on either type. Topological maps can be transformed in anyway and still have the same meaning whereas metric maps are limited in the number of transform that keep the map coherent and meaningful to human viewers. Below we describe common map types that have developed from these two categories, most of which are metric maps or a hybrids of the two. Figure \hyperref[fig:MapRep]{1} shows a visual of a few of these representation.

\subsection{Topological Map}
Topological map representations are essentially graph based structures where distinct locations are nodes connected by edges that define traversable paths. Distinct places are locations with enough memorable features in the environment such that localization is easily achievable. Common methods to construct topological maps derive from Voronoi diagrams\cite{aurenhammer1991voronoi}, which can be created incrementally, as the robot explores, or from a given metric map\cite{choset1996sensor,filliat2003map,thrun1998learning}. This form of representations is extremely light and good for goal based planning and navigation task.


\subsection{Polygonal Map}
When representing an environment as a polygonal region, the core idea is that our sensory data (primarily laser range scans or sonar readings), can be transformed into geometric primitives such as lines and spline curves. There are a variety of methods to transform the data to primitives and stitch together these primitives to create polylines\cite{veeck28learning,paskin2012robotic, gonzalez2002navigation}. These polylines are then used to make a polygonal representation of the environment. 
The advantages of this representation is it reduces discretization errors and is compact.  

\subsection{Occupancy Grids}
Occupancy grids have enjoyed enormous popularity and stand as one of the most used map representations. Part of the reason for this is how easily the map can be updated from sensor data (primarily laser scans). Occupancy grids represent the environment as discretized two-dimensional grid cells or three-dimensional voxels. Cells are given values that represent belief in occupancy(free or occupied). This is a probabilistic map with a common assumption that the grid cells are independent of each other. Although this is hardly ever true, it allows us to easily use the log-odds to update the posterior probability of each cell with negligible error for most environments \cite{thrun2002robotic,filliat2003map}. This type of representation has a large memory requirements and can have discretization errors. However, it is possible to discretize the space irregularly in order to save memory. The robustness and easy of implementation are two reasons this representation is heavily used.

\subsection{Feature Map}
Feature based maps create a metric map of the environment using common landmarks such as points and objects; walls and other distinguishable features are also possible.
However, assumptions must now be made on what type of features we expect to identify in the region \cite{wurm2010bridging,filliat2003map}. This form of representation is much lighter than occupancy grid maps and can still provide a layout of the environment, that is close to peoples perceptions of environments \cite{thrun2002robotic}.

\subsection{Semantic Map} 
Semantic map representations are similar to topological representations in that they represent the environment though place-based representations. This notion can be extended from just locations and paired with metric models mentioned above. The key idea is that semantic maps should represent high-level features with labels corresponding to humans-concepts (kitchen,desk,hallway); as well as their relations to each other. See \cite{kostavelis2015semantic} for more details.

\begin{figure} [!ht]
\includegraphics[width=0.48\textwidth]{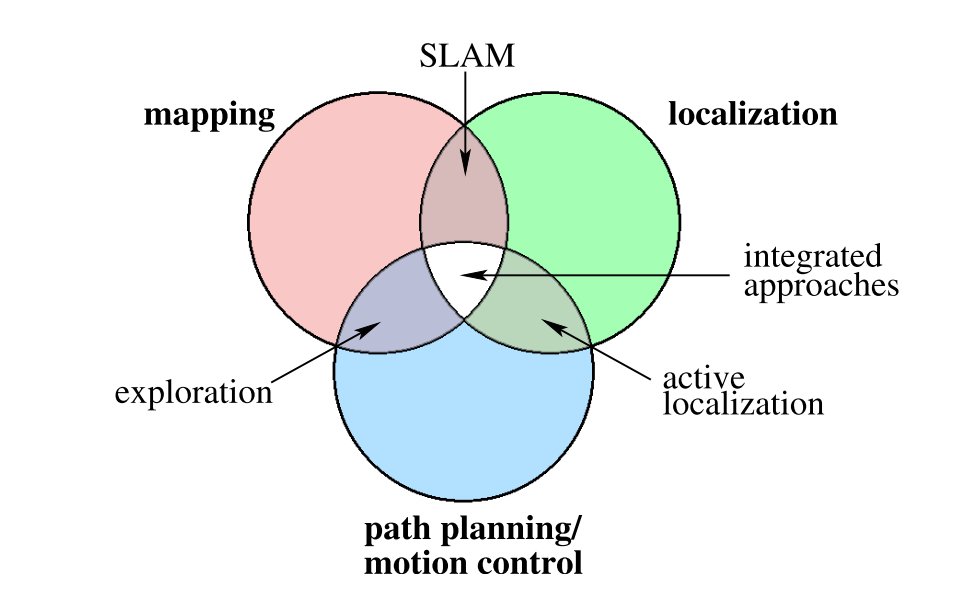}
\label{fig:VinDig}
\caption{The figure above is adapted from \cite{stachniss2009robotic} and shows a Venn digraph of the task a robot needs to solve in order to accurately create a model of an environment. Exploration can be seen in the intersection of path planning and mapping.}
\end{figure}

\section{Exploration methods}
\label{sec:ExpMethod}
The goal of all exploration methods is to explore an unknown environment as quickly and accurately as possible. The \textbf{central question} that each method tries to answer is: \textit{given what you know about the world, where should you move next to gain as much information a possible} \cite{yamauchi1997frontier}. Here information depends on the objective of the robot. In this paper, we frame exploration primarily in the context of mapping - were we want a full model of the environment. Nonetheless, the methods presented can and have been applied to other exploration context: search and rescue, environmental monitoring, planetary science, and oceanography.\cite{dunbabin2012robots,low2007adaptive,dunbabin2016autonomous} 

Regardless of the objective, when thinking of an exploration algorithm, it is important to keep in mind the representation that will be used since nearly all methods depend on the representation to achieve their goal. The interdependence between exploration algorithms and representations creates significant overlap between closely related fields (Figure \hyperref[fig:VinDig]{2}). In path planning, the environment and robot position is fully or partially known and the objective is to find an efficient path to a specific location. SLAM deals with how to accurately create a model of the environment and localize the robot within the environment correctly. Exploration generally assumes the pose of the robot is known or can be identified and instead abstracts focus to guiding the robot through the whole environment. Active SLAM and other integrated approaches, are some combinations of these. The rest of this section will introduce methods presented in various papers, relating to the three methods formalized in section \hyperref[sec:Problem Formulation]{II}


\subsection{Greedy Based Exploration}
The greedy technique attempt to answer the central question of where to move next is rather simple: move to the closest reachable position, with the highest utility, that is on the frontier or outside of what is know but presumably, in a safe and reachable location. The primary implementation of these methods are frontier and sampling based, as well as other myopic approaches. A frontier is a region on the boundary between what is known and unknown, it was first introduced by Brian Yamauchi \cite{yamauchi1997frontier}. Frontiers are created and grouped into clusters depending on the implementation. This concept has been improved upon and widely used within exploration task \cite{campos2017complete,rekleitis2001multi,meghjani2012multi,yamauchi1998frontier,holz2010evaluating,choudhary2018autonomous}. Even earlier and also influential was Kuipers and Byan's work on Spatial Semantic Hierarchy\cite{kuipers1991robot}. They highlight the notion that exploration is not only about building an accurate map but performing spatial learning; showing how a hybrid topological and geometric map can be used to do both. The topological map gives precedent to exploration and semantic spatial understanding, while also allowing distinct places (nodes) and paths to accumulate local metric information. Their exploration is based on the topology: once at a distinct place, move into an open direction and keep track of each distinct place and the directions not yet explored. Howie Choset's et al. \cite{choset1996sensor} work implements a similar approach in the real world. They first explain how a topological graph can be created in an unknown environment using generalized Voronoi graph. Like Kuipers, their exploration is akin to graph search. A follow up of this work is shown in \cite{nagatani1999toward}. 

Recent work by \cite{campos2017complete} uses some of the foundational methods above and maintains a hybrid map that contains local metric maps and a topology of connections created from their own version of a Voronoi-like graph. They alters frontiers based exploration to explicit preform loop closer that creates better maps by essentially ensuring frontiers are revisited. A comparison of a few other frontier and information based methods is shown \cite{holz2010evaluating}. They also improve the vanilla frontier method in two ways. First by repetitively re-checking if the frontier being approached is still a valid frontier, selecting the next closet frontier if not. Then by segmenting areas mapped based on distinct places; giving preference to frontiers in the same segment as the robot. This prevents situations where the robot may leave a room half explored while moving to the next closest frontier. Another approach, shown in \cite{thrun2004autonomous}, autonomously explore an abandoned mine by setting consecutive goal locations in the desired travel direction (an average previous robot positions). Their environment is simplified compared to others since the mine is primarily straight with a right bend. 

There has also be quite a bit of work on multi-robot greedy exploration. The addition of other robots can greatly decrease exploration time and to do so efficiently, most work focuses on the coordination between the robots. This is shown in work by Matthew Dunbabin \cite{dunbabin2016autonomous} where four autonomous surface vehicles work together to explore water reservoirs in order to detect methane "hot-spots". The environment is partially known but the locations of the needed "hot-spots" is not. They select sample locations using a random walk and potential fields. All robots communicate and represent previously sample sites as a 2D Gaussian; a random position at a set radius from each robots current position is selected and chosen to be sampled if not on land, and the value of the the closest Gaussian potential is below a threshold. The set radius changes based on sampling measures. Their method allows the robots to individually explore locations efficiently, while not overlapping with other robots. \cite{rekleitis2001multi} uses multiple agents to explore a large spaces and focus primarily on the accuracy of the map. Their exploration technique uses trapezoidal planar decomposition to segment and systematically sweep through an area; making sure the static robot is always in view of the moving robot. This allows for better localization to eliminate drift. This algorithm was expanded and implemented from their previous paper \cite{rekleitis1997multi} 

Low et al. \cite{low2007adaptive} presents an adaptive cluster sampling exploration approach that reduces mission time and energy consumption compared to other sampling approaches. Similar to \cite{dunbabin2016autonomous} they are interested in detecting "hotspots", or locations of interest, rather than complete coverage. Their approach first performs simple random sampling universally around the robot, then samples more thoroughly the neighborhoods with information above a threshold. All robots communicate new possible sampling locations with each other and \cite{lagoudakis2005auction} is used for coordination of the robots. The work in \cite{meghjani2012multi} considers scenarios where robots start with unknown locations and no communication while exploring. The objective is to individually explore while trying to meet up and share information. They use a frontier based exploration but keep track of distinct places they can navigate back to for rendezvous. 


Through the examples listed, it is clear that there are many approaches that do not strictly follow the formalization in section \hyperref[sec:Greddy]{II.C}. Nonetheless, the approaches presented achieve this greedy optimization inherently and are myopic.  

\subsection{Information Gain Based Exploration}
\label{sec:InfoGain}
In this section we discuss work on exploration that is driven by a particular form of information and is non-myopic. We use the term non-myopic loosely because although some methods have a short planning horizon, they make their decisions on where to move next based on their reasoning about the expected future.   

Atanasov et al. recent work on information acquisition \cite{atanasov2014information} presents a truly non-myopic algorithm with performance guarantees and theoretical error bounds. Their focus is to optimize the trajectory of the robot, as it explores, in order to create more accurate maps. They show that this problem can be can be reduced from a stochastic optimal control problem to deterministic optimal control of which open loop policies are better suited - this reduces the state space of the sensors and allows offline computations. Using a forward value iteration approach, a search ‘tree of trajectories’ is created to represent the expected reachable states given a set of controls. Crossing paths or sensing from similar configurations, are pruned to remove less informative paths. The optimal trajectory is selected as the path which maximizes the mutual information of the sensors and map estimate. \cite{sim2005global} also priorities the accuracy of the map while exploring and similarly focuses on optimizing the trajectory of the robot. They discretize the pose and action space of the robot according to the gird representation of the environment. Then using breath first search over all poses up to a threshold, and pruning repeated states, they search for the expected map estimates that leads to maximum information gain. The paper \cite{bourgault2002information} came slightly before the other two and presents a similar approach where they weigh the best trajectory depending upon the expected information gain of the map and robot localization accuracy. The notion being that there is little value in exploring when the robot is not accurately localized. The work of Cyrill Stachniss \cite{stachniss2009robotic} presents a detailed formalization of entropy based information gain approaches, as well as more example using multiple robots and cases were the pose of the robot is not assumed to be known.  

Exploration based on the expected next-best view of the robot is shown in \cite{gonzalez2002navigation, bircher2016receding}. In \cite{gonzalez2002navigation} they define the concept of safe regions as: the largest region guaranteed to be free of objects based on a constrained set of frontiers. Candidate positions of where to move next are randomly selected within safe regions and in sight of the constrained frontiers. The best position is determined by a combination of potential visibility gain and distance to the position. Visibility gain is estimated based on sensor range and expected overlap of previous mapped regions. This proves to be effective for indoor environments. \cite{bircher2016receding} similarly computes where to move next based on distance and expected visibility of the location. They use RRT \cite{lavalle1998rapidly} in the configuration space, to create a tree of vehicle configurations up to set length; where edges represents paths between configurations. They discrete the world into voxels and associate each configuration with a set of voxels marked visible or unmapped based on the sensor data and confidence level. Finally, they consider the quality of information gained as the summation of unmapped voxels that can be explored a long a branch of the tree, penalized by the distance to the node.

In section \hyperref[sec:InfoGain]{II.B} we formalize information gain using an entropy based approach. While this is true of some of the methods discussed above, it is not fully representative of methods such as "next-best view" and other more ad hoc methods with tailored utility functions. A full formulation of next-best view methods can be found in \cite{zhu2018deep}. The key reason for grouping these methods within this section is that: similar to the entropy based approach and formalization, the other exploration methods in this section are characterized by an objective function driven on the impact to future states estimates rather than optimizing current state.

\subsection{Learning Based Exploration}
Learning based methods for exploration can be consider the most resent approach of the three methods presented. This method is primarily data driven; the idea being that as robots explore a space, a policy of how to efficiently explore can be learned and used to explore similar environments. 

Kollar and Roy highlight this method in their paper \cite{kollar2008trajectory}, where they use a reinforcement learning (RL) based approach to optimize the trajectory of a robot exploring an office space. Their objective function prioritize exploration as a balance between coverage and accuracy. Rather than simply summing each weighed term, they define coverage as choosing \textit{n} locations that maximize the likelihood of complete environment observation, then constrain their trajectory optimization for accuracy by these location. This is essentially two optimization problems and quickly becomes intractable to solve without a prior known map. It is here that they turn to RL, using Policy Search by Dynamic Programming as their learning algorithm. The show that the robot is able to learn a policy that creates a more accurate map than a shortest-path algorithm both in simulation and the real-world. We refer readers to the paper for more implementation details.

In \cite{burgard2000collaborative} they implement a kind of next-best view approach, while also ensuring multi-robot collaboration. Their focus is to minimize the time needed for complete map exploration. They use frontiers to identify potential target points and value iteration to assign a cost to each target. The cost of a target is equal to the global probability that the location (cell) is occupied, multiplied by the local distance to the location. This is then computed for neighboring cells and included as part of the current cells cost. While exploring the robots keep a histogram of the distances covered by their sensors. This is used to estimate the expected visibility range of each robot. Initially the utility of all target locations are equal, however, once a robot selects a target location, the utility of adjacent locations within expected view are lowered. The best target point is selected as a trade off of the cost and utility. This selection is similar to a greedy approach but the utility is dependent on the history of exploration. \cite{richter2016learning} also performs a learning based approach that incorporates next-bast view. They begin by explaining that exploration methods often involve selecting a navigation goal and their objective is to create a heuristic for selecting actions based on the cost-to-goal after an action. They define cost as the time duration of an action and utility as the estimated remaining time to reach the goal afterwards. This of course is unknown and must be estimated. Using a data set of labeled belief-action pairs, they are able to train a model to learn which actions make the most progress towards the goal(defined by distance). They use this model and a greedy shortest path method, using the robots motion dynamics, as the utility in their final heuristic; selecting actions that minimize their heuristic. The scope of their work is limited to corners in hallway environments, through which they show that the learned model helps chose actions that give greater visibility when rounding a corner. This allows the robot to safely explore with greater speed.

All of the approaches presented in this section explore indoor environments. While it is unclear if a learning based approach would be able to draw insights in unstructured environments such as the outdoors, this is an open area of work. 

\begin{table*}[t]
  \label{sec:Table1}
  \centering
  \caption{} 
  \begin{tabular}{|l|ccc|}
    \hline
     \bf Map Representation & \bf Greedy & \bf Information Gain & \bf Learning \\ 
    \hline
    \hline
     Topological-Metric Hybrid & \cite{dunbabin2016autonomous,kuipers1991robot,choset1996sensor,nagatani1999toward,rekleitis2001multi}  & \cite{gonzalez2002navigation} & \cite{DBLP:journals/aim/NourbakhshPB95} \\ &  \cite{meghjani2012multi,campos2017complete,yan2010sampling,wurm2008coordinated} & &  \\
    \hline
    Occupancy Grid &  \cite{yamauchi1997frontier,yamauchi1998frontier,low2007adaptive} & \cite{sim2005global, atanasov2014information,stachniss2006speeding} & \cite{kollar2008trajectory,burgard2000collaborative,richter2016learning}\\ & \cite{holz2010evaluating,wurm2008coordinated} & \cite{bourgault2002information,bircher2016receding} &\cite{zhu2018deep,richter2017autonomous} \\
    \hline 
    Semantic & \cite{kuipers1991robot} & \cite{stachniss2006speeding} &\\
    \hline
    Feature & \cite{thrun2004autonomous,holz2010evaluating}  & \cite{bircher2016receding} & \cite{kollar2008trajectory}\\
    \hline
  \end{tabular}
  \\[10pt]
  \caption*{The table above categories the research literature in this survey according to their respective exploration method and map representation used while exploring or presented as a final map model.}
\end{table*}

\section{Exploration: Particular use case}
\label{sec:ExpPC}
\hyperref[sec:Table1]{Table I} categorizes some of the work reviewed in this paper based on map representation and exploration methodology. It is important to note that the categorization is based primarily on the representation used while exploring, as well as additional map representations created as the final model of the environment. In this section we give a meta overview of what methods tend to be used and why - in relation to representations and objectives.

For greedy exploration methods, we see a high concentration of work using topological and metric map hybrids. Many of these methods incorporate Voronoi graphs, frontier, or sampling techniques that are easily paired to create topological graph based maps. Essentially, topological maps are created for exploration while metric maps are created as a product of this exploration; or coupled as hybrids with local metric maps connected to each other. This is attractive for exploration where rigid geometric accuracy is prioritized less than minimizing time to complete coverage.

Overwhelmingly, the greedy approach is utilized more than the other two approaches. Despite having little to no performance guaranties, the ease of implementation for these methods make them greatly desirable. In addition to this, the fact that when compared to the other methods, they tend to do just as well or better in some cases and if not only marginally worse (particularly in terms of time), adds to their popularity \cite{amigoni2008experimental, holz2010evaluating}. For those willing and in need of accuracy or efficiently in terms of energy consumption due to sensor measurements and vehicle path length, information gain and learning based approaches are more optimal choices. 

In terms of map representation used for exploration, occupancy grids - unsurprisingly do to their popularity, are found heavily in all three approaches. As stated before, part of their popularity lies in the ease of implementation and seemingly natural pairing with common sensors used for exploration (lasers). In fact, when it comes to information gain and learning based approaches, occupancy grids make up practically all representations used. In contrast, feature maps such as point clouds, are used to provide representations to human viewers but are hardly used in conjunction with exploration algorithm. Similarly, there is a significant lack of semantic map representation used for exploration. This is in part due to the fact that we are only now on the cusp of being able to identify whole scene locations reliably.  

\section{Discussion: Open Problems} 
\label{sec:DiscussionOP}
Work on robotic exploration has been fundamental towards the fields of autonomous navigation and mapping. The impact of this has been seen in other disciplines such as oceanography, planetary science, and disaster relief. In this sections we briefly discuss two areas where we see room for more studies in the field of exploration. 

The first area is within learning based approaches. These methods are still maturing and as a result have much fewer implementations than others. The truly novel and exciting difference in these approaches are their ability to learn and improve their objective function from experience. This creates an opportunity to learn exploration techniques that can be generalized to similar environments, like frontier based approaches, while still remaining optimized for an environment. This is possible because although each environment has a distinct layout, there are often similarities within the latent spatial arrangement that experience can highlight. It would be interesting to see this approach groomed such that a robot could learn the type of environment it is in and set an appropriate objective function to explore efficiently. 

The second area we see room for more research, has to do with semantic representation used in exploration. As noted in this survey, there is not a lot of work in this field due to previous limitations in semantic location classification. Now that we are more accurately able to identify location, similarly to learning based approaches, methods that take advantage of learning semantic locations can leverage the similarities within latent spatial arrangement that indoor environments tend to have. This kind of work can be effective towards faster exploration to locations of interest and full coverage.

From the discussion, we highlight an acute observation. When looking at the spectrum of environments and exploration methods, we note that unstructured environments such as the outdoors, tend to use greedy based approaches like sampling; whereas more structured environments (homes and offices) tend to use a variety of approaches but almost exclusively find information gain and learning based approaches only used within this context. The areas we propose for more research fall best within structured environments.

\section{Conclusion}
\label{sec:conclusion}
We present a survey of the exploration methods used within robotic literature as it pertains to field robotics; particularly framed for mapping approaches but relevant to other objectives. 
Although we try to categorize the methods for exploration as efficiently and distinctly as possible, many methods draw inspiration from each other and are combination of approaches more than a single approach. These ad hoc objective functions tailored to specific objectives can make classification seem slightly frivolous. Thought this survey, however, we show that although the methods may overlap, their motivation and the overarching approach used are distinct and impact how the space is explored.

\section*{Acknowledgment}
This material is based upon work supported by the National Science Foundation Graduate
Research Fellowship Program under Grant No. DGE-1650112. Any opinions,
findings, and conclusions or recommendations expressed in this material are those of the author(s) and do not necessarily reflect the views of the National Science Foundation.

\nocite{*}
\bibliographystyle{plain}
\bibliography{explore_refs}

\begin{thebibliography}{10}

\bibitem{amigoni2008experimental}
Francesco Amigoni.
\newblock Experimental evaluation of some exploration strategies for mobile
  robots.
\newblock In {\em 2008 IEEE International Conference on Robotics and
  Automation}, pages 2818--2823. IEEE, 2008.

\bibitem{atanasov2014information}
Nikolay Atanasov, Jerome Le~Ny, Kostas Daniilidis, and George~J Pappas.
\newblock Information acquisition with sensing robots: Algorithms and error
  bounds.
\newblock In {\em 2014 IEEE International Conference on Robotics and Automation
  (ICRA)}, pages 6447--6454. IEEE, 2014.

\bibitem{aurenhammer1991voronoi}
Franz Aurenhammer.
\newblock Voronoi diagrams—a survey of a fundamental geometric data
  structure.
\newblock {\em ACM Computing Surveys (CSUR)}, 23(3):345--405, 1991.

\bibitem{bailey2006simultaneous}
Tim Bailey and Hugh Durrant-Whyte.
\newblock Simultaneous localization and mapping (slam): Part {II}.
\newblock {\em IEEE Robotics \& Automation Magazine}, 13(3):108--117, 2006.

\bibitem{bircher2016receding}
Andreas Bircher, Mina Kamel, Kostas Alexis, Helen Oleynikova, and Roland
  Siegwart.
\newblock Receding horizon" next-best-view" planner for 3d exploration.
\newblock In {\em 2016 IEEE international conference on robotics and automation
  (ICRA)}, pages 1462--1468. IEEE, 2016.

\bibitem{bourgault2002information}
Frederic Bourgault, Alexei~A Makarenko, Stefan~B Williams, Ben Grocholsky, and
  Hugh~F Durrant-Whyte.
\newblock Information based adaptive robotic exploration.
\newblock In {\em IEEE/RSJ international conference on intelligent robots and
  systems}, volume~1, pages 540--545. IEEE, 2002.

\bibitem{burgard2000collaborative}
Wolfram Burgard, Mark Moors, Dieter Fox, Reid Simmons, and Sebastian Thrun.
\newblock Collaborative multi-robot exploration.
\newblock In {\em ICRA}, pages 476--481, 2000.

\bibitem{cadena2016past}
Cesar Cadena, Luca Carlone, Henry Carrillo, Yasir Latif, Davide Scaramuzza,
  Jos{\'e} Neira, Ian Reid, and John~J Leonard.
\newblock Past, present, and future of simultaneous localization and mapping:
  Toward the robust-perception age.
\newblock {\em IEEE Transactions on robotics}, 32(6):1309--1332, 2016.

\bibitem{campos2017complete}
Francisco~M Campos, Miguel Marques, Fernando Carreira, and JMF Calado.
\newblock A complete frontier-based exploration method for pose-slam.
\newblock In {\em 2017 IEEE International Conference on Autonomous Robot
  Systems and Competitions (ICARSC)}, pages 79--84. IEEE, 2017.

\bibitem{choset1996sensor}
Howie~M Choset and Joel Burdick.
\newblock Sensor based motion planning: The hierarchical generalized voronoi
  graph.
\newblock 1996.

\bibitem{choudhary2018autonomous}
Abhishek Choudhary.
\newblock Autonomous exploration and data gathering with a drone, 2018.

\bibitem{dunbabin2016autonomous}
Matthew Dunbabin.
\newblock Autonomous greenhouse gas sampling using multiple robotic boats.
\newblock In {\em Field and Service Robotics}, pages 17--30. Springer, 2016.

\bibitem{dunbabin2012robots}
Matthew Dunbabin and Lino Marques.
\newblock Robots for environmental monitoring: Significant advancements and
  applications.
\newblock {\em IEEE Robotics \& Automation Magazine}, 19(1):24--39, 2012.

\bibitem{durrant2006simultaneous}
Hugh Durrant-Whyte and Tim Bailey.
\newblock Simultaneous localization and mapping: part {I}.
\newblock {\em IEEE Robotics \& Automation Magazine}, 13(2):99--110, 2006.

\bibitem{filliat2003map}
David Filliat and Jean-Arcady Meyer.
\newblock Map-based navigation in mobile robots:: I. a review of localization
  strategies.
\newblock {\em Cognitive Systems Research}, 4(4):243--282, 2003.

\bibitem{gonzalez2002navigation}
H{\'e}ctor~H Gonz{\'a}lez-Banos and Jean-Claude Latombe.
\newblock Navigation strategies for exploring indoor environments.
\newblock {\em The International Journal of Robotics Research},
  21(10-11):829--848, 2002.

\bibitem{holz2010evaluating}
Dirk Holz, Nicola Basilico, Francesco Amigoni, and Sven Behnke.
\newblock Evaluating the efficiency of frontier-based exploration strategies.
\newblock In {\em ISR 2010 (41st International Symposium on Robotics) and
  ROBOTIK 2010 (6th German Conference on Robotics)}, pages 1--8. VDE, 2010.

\bibitem{kaelbling1998planning}
Leslie~Pack Kaelbling, Michael~L Littman, and Anthony~R Cassandra.
\newblock Planning and acting in partially observable stochastic domains.
\newblock {\em Artificial intelligence}, 101(1-2):99--134, 1998.

\bibitem{kollar2008trajectory}
Thomas Kollar and Nicholas Roy.
\newblock Trajectory optimization using reinforcement learning for map
  exploration.
\newblock {\em The International Journal of Robotics Research}, 27(2):175--196,
  2008.

\bibitem{kostavelis2015semantic}
Ioannis Kostavelis and Antonios Gasteratos.
\newblock Semantic mapping for mobile robotics tasks: A survey.
\newblock {\em Robotics and Autonomous Systems}, 66:86--103, 2015.

\bibitem{kuipers1991robot}
Benjamin Kuipers and Yung-Tai Byun.
\newblock A robot exploration and mapping strategy based on a semantic
  hierarchy of spatial representations.
\newblock {\em Robotics and Autonomous systems}, 8(1-2):47--63, 1991.

\bibitem{lagoudakis2005auction}
Michail~G Lagoudakis, Evangelos Markakis, David Kempe, Pinar Keskinocak,
  Anton~J Kleywegt, Sven Koenig, Craig~A Tovey, Adam Meyerson, and Sonal Jain.
\newblock Auction-based multi-robot routing.
\newblock In {\em Robotics: Science and Systems}, volume~5, pages 343--350.
  Rome, Italy, 2005.

\bibitem{lavalle1998rapidly}
Steven~M LaValle.
\newblock Rapidly-exploring random trees: A new tool for path planning.
\newblock 1998.

\bibitem{low2007adaptive}
Kian~Hsiang Low, Geoffrey~J Gordon, John~M Dolan, and Pradeep Khosla.
\newblock Adaptive sampling for multi-robot wide-area exploration.
\newblock In {\em Proceedings 2007 IEEE International Conference on Robotics
  and Automation}, pages 755--760. IEEE, 2007.

\bibitem{martinez2009bayesian}
Ruben Martinez-Cantin, Nando de~Freitas, Eric Brochu, Jos{\'e} Castellanos, and
  Arnaud Doucet.
\newblock A bayesian exploration-exploitation approach for optimal online
  sensing and planning with a visually guided mobile robot.
\newblock {\em Autonomous Robots}, 27(2):93--103, 2009.

\bibitem{meghjani2012multi}
Malika Meghjani and Gregory Dudek.
\newblock Multi-robot exploration and rendezvous on graphs.
\newblock In {\em 2012 IEEE/RSJ International Conference on Intelligent Robots
  and Systems}, pages 5270--5276. IEEE, 2012.

\bibitem{nagatani1999toward}
Keiji Nagatani and Howie Choset.
\newblock Toward robust sensor based exploration by constructing reduced
  generalized voronoi graph.
\newblock In {\em Proceedings 1999 IEEE/RSJ International Conference on
  Intelligent Robots and Systems. Human and Environment Friendly Robots with
  High Intelligence and Emotional Quotients (Cat. No. 99CH36289)}, volume~3,
  pages 1687--1692. IEEE, 1999.

\bibitem{DBLP:journals/aim/NourbakhshPB95}
Illah~R. Nourbakhsh, Rob Powers, and Stan Birchfield.
\newblock {DERVISH} - an office-navigating robot.
\newblock {\em {AI} Magazine}, 16(2):53--60, 1995.

\bibitem{paskin2012robotic}
Mark Paskin and Sebastian Thrun.
\newblock Robotic mapping with polygonal random fields.
\newblock {\em arXiv preprint arXiv:1207.1399}, 2012.

\bibitem{platt2010belief}
Robert Platt~Jr, Russ Tedrake, Leslie Kaelbling, and Tomas Lozano-Perez.
\newblock Belief space planning assuming maximum likelihood observations.
\newblock 2010.

\bibitem{rekleitis2001multi}
Ioannis Rekleitis, Gregory Dudek, and Evangelos Milios.
\newblock Multi-robot collaboration for robust exploration.
\newblock {\em Annals of Mathematics and Artificial Intelligence},
  31(1-4):7--40, 2001.

\bibitem{rekleitis1997multi}
Ioannis~M Rekleitis, Gregory Dudek, and Evangelos~E Milios.
\newblock Multi-robot exploration of an unknown environment, efficiently
  reducing the odometry error.
\newblock In {\em IJCAI}, volume~2, pages 1340--1345, 1997.

\bibitem{richter2016learning}
Charles Richter and Nicholas Roy.
\newblock Learning to plan for visibility in navigation of unknown
  environments.
\newblock In {\em International Symposium on Experimental Robotics}, pages
  387--398. Springer, 2016.

\bibitem{richter2017autonomous}
Charles~Andrew Richter.
\newblock {\em Autonomous navigation in unknown environments using machine
  learning}.
\newblock PhD thesis, Massachusetts Institute of Technology, 2017.

\bibitem{sim2005global}
Robert Sim and Nicholas Roy.
\newblock Global a-optimal robot exploration in slam.
\newblock In {\em Proceedings of the 2005 IEEE international conference on
  robotics and automation}, pages 661--666. IEEE, 2005.

\bibitem{stachniss2009robotic}
Cyrill Stachniss.
\newblock {\em Robotic mapping and exploration}, volume~55.
\newblock Springer, 2009.

\bibitem{stachniss2005information}
Cyrill Stachniss, Giorgio Grisetti, and Wolfram Burgard.
\newblock Information gain-based exploration using rao-blackwellized particle
  filters.

\bibitem{stachniss2006speeding}
Cyrill Stachniss, O~Martinez Mozos, and Wolfram Burgard.
\newblock Speeding-up multi-robot exploration by considering semantic place
  information.
\newblock In {\em Proceedings 2006 IEEE International Conference on Robotics
  and Automation, 2006. ICRA 2006.}, pages 1692--1697. IEEE, 2006.

\bibitem{thrun1998learning}
Sebastian Thrun.
\newblock Learning metric-topological maps for indoor mobile robot navigation.
\newblock {\em Artificial Intelligence}, 99(1):21--71, 1998.

\bibitem{thrun2002robotic}
Sebastian Thrun.
\newblock Robotic mapping: A survey.
\newblock {\em Exploring artificial intelligence in the new millennium},
  1(1-35):1, 2002.

\bibitem{thrun2005probabilistic}
Sebastian Thrun, Wolfram Burgard, and Dieter Fox.
\newblock {\em Probabilistic robotics}.
\newblock MIT press, 2005.

\bibitem{thrun2004autonomous}
Sebastian Thrun, Scott Thayer, William Whittaker, Christopher Baker, Wolfram
  Burgard, David Ferguson, Dirk Hahnel, D~Montemerlo, Aaron Morris, Zachary
  Omohundro, et~al.
\newblock Autonomous exploration and mapping of abandoned mines.
\newblock {\em IEEE Robotics \& Automation Magazine}, 11(4):79--91, 2004.

\bibitem{veeck28learning}
M~Veeck and W~Burgard.
\newblock Learning polyline maps from range scan data acquired with mobile
  robots.
\newblock In {\em Proceedings. IEEE/RSJ International Conference on Intelligent
  Robots and Systems (IROS 2004)}, volume~28, pages 1065--1070.

\bibitem{wurm2008coordinated}
Kai~M Wurm, Cyrill Stachniss, and Wolfram Burgard.
\newblock Coordinated multi-robot exploration using a segmentation of the
  environment.
\newblock In {\em 2008 IEEE/RSJ International Conference on Intelligent Robots
  and Systems}, pages 1160--1165. IEEE, 2008.

\bibitem{wurm2010bridging}
Kai~M Wurm, Cyrill Stachniss, and Giorgio Grisetti.
\newblock Bridging the gap between feature-and grid-based slam.
\newblock {\em Robotics and Autonomous Systems}, 58(2):140--148, 2010.

\bibitem{yamauchi1997frontier}
Brian Yamauchi.
\newblock A frontier-based approach for autonomous exploration.
\newblock {\em IEEE International Symposium on Computational Intelligence in
  Robotics and Automation}, 97:146, 1997.

\bibitem{yamauchi1998frontier}
Brian Yamauchi et~al.
\newblock Frontier-based exploration using multiple robots.
\newblock In {\em Agents}, volume~98, pages 47--53, 1998.

\bibitem{yan2010sampling}
Zhi Yan, Nicolas Jouandeau, and Arab~Ali Cherif.
\newblock Sampling-based multi-robot exploration.
\newblock In {\em ISR 2010 (41st International Symposium on Robotics) and
  ROBOTIK 2010 (6th German Conference on Robotics)}, pages 1--6. VDE, 2010.

\bibitem{zhu2018deep}
Delong Zhu, Tingguang Li, Danny Ho, Chaoqun Wang, and Max Q-H Meng.
\newblock Deep reinforcement learning supervised autonomous exploration in
  office environments.
\newblock In {\em 2018 IEEE International Conference on Robotics and Automation
  (ICRA)}, pages 7548--7555. IEEE, 2018.

\end{thebibliography}

\end{document}